\documentclass{cys}

\usepackage[utf8]{inputenc}
\usepackage{tipa}

\usepackage[english]{babel}

\usepackage{graphicx}

\usepackage{epstopdf} 

\addto\captionsenglish{%

}

\addto\captionsspanish{%

}

\title{Part-of-speech tagging for Nagamese Language using CRF}

\author{Alovi N Shohe$^1$, Chonglio Khiamungam$^1$, Teisovi Angami*$^1$\\}

\affil{ 
$^1$ Department of Information Technology,\\
        Nagaland University, Kohima Campus, India \authorcr   
           
\authorcr \authorcr \authorcr
nalovishohe@gmail.com, khiamniunganchongliu@gmail.com, teisovi@nagalanduniversity.ac.in
}

\begin{document}

\maketitle

\renewcommand{\tablename}{Table}

\begin{abstract}
This paper investigates part-of-speech tagging, an important task in Natural Language Processing (NLP) for the Nagamese language. The Nagamese language, a.k.a. Naga Pidgin, is an Assamese-lexified Creole language developed primarily as a means of communication in trade between the Nagas and people from Assam in northeast India. A substantial amount of work in part-of-speech-tagging has been done for resource-rich languages like English, Hindi, etc. However, no work has been done in the Nagamese language. To the best of our knowledge, this is the first attempt at part-of-speech tagging for the Nagamese Language. The aim of this work is to identify the part-of-speech for a given sentence in the Nagamese language. An annotated corpus of 16,112 tokens is created and applied machine learning technique known as Conditional Random Fields (CRF). Using CRF, an overall tagging accuracy of 85.70\%; precision, recall of 86\%, and f1-score of 85\% is achieved.
\end{abstract}


\section{Introduction}
Nagamese (Naga Pidgin) is an important Creole language of Nagaland located in the North-Eastern part of India, apart from the tribal languages spoken, it is used as a common language across the entire state of Nagaland. It is an Assamese-lexified (Assam is an Indian state bordering Nagaland) creole language developed primarily as a means of communication in trade between the Nagas and people from Assam. It is widely used in mass media in the news, radio stations, state-government media, etc. The Nagamese language is a resource-poor language, and therefore, it is a challenge to create resources for applying various language processing tasks.

Part-of-speech (POS) tagging involves labeling each word in a sentence with its appropriate part of speech.

Example: Itu/ADJECTIVE dikhikena/VERB Isor/NOUN khusi/ADJECTIVE lagise/VERB ./SYM (God was pleased with what He saw.)

It is not an easy task due to the dependence of POS tags on contextual information.

In this work, a POS tagger is built for the Nagamese Language using Conditional Random Fields (CRF) which is a machine-learning technique. The main contribution of our work is the identification of the POS tagset, the creation of an annotated corpus of 16,115 tokens, and its evaluation using CRF. A discussion on the error analysis of the tagging performance is also presented.

The rest of the paper is organized as follows: Section~\ref{section:the_nagamese_language} gives an introduction to the Nagamese Language, Section~\ref{section:related_works} gives an overview of the related works, Section~\ref{section:pos_tagset} gives a description of the POS tagset, Section~\ref{section:methodology} discusses the methodology, Section~\ref{section: experimental_results} reports and discusses the experimental results and Section~\ref{section:conclusion} draws the conclusion and discusses future works.

\section{The Nagamese Language} \label{section:the_nagamese_language}
This section provides an overview of the character set, syllabic pattern, and grammar for the Nagamese Language.

\subsection{Character Set for Nagamese Language}
The Nagamese language has 28 phonemes, including 6 vowels and 22 consonants.
\begin{description}
\item[Vowels:] i, u, e, \textschwa, o, a
\item[Consonants:]  \( p, t, c, k, b, d, j, g, p^{h},t^{h}, c^{h}, k^{h}, m, n,\dot{n},\) \( s,\check{s}, h, r, I, w, y \)
    
\end{description}

A sentence in Nagamese is given below:
\begin{center}"Moy dos baje pora yeti ase." \newline
(I am waiting here from 10 o \rq clock.)\end{center}

\subsection{Syllabic Pattern}
As found in the work of Sreedhar \cite{sreedhar1985standardized}, a word in the Nagamese language may consist of one or more syllables ranging up to a maximum of four syllables. The entire monosyllabic words in this language could be sub-grouped into six classes, which when put in a schematic formula would be: 
\begin{center}$(C) (C) V (C) (C)^{2}$ \end{center}

The only limitation in the operation of the above formula is that V cannot occur alone. A disyllabic word in the Nagamese language cannot consist of just two vowels alone. The structure of the disyllabic words in this language can be broadly sub-grouped into two which is shown here:

\begin{small}
\begin{enumerate}
\item $V(C)(C)(C)V(C)$
\item $(C)CV(C)(C)CV(C)(C)\ or\ (C)CV(C)(C)V(C)(C)$ 
\end{enumerate}
\end{small}

The trisyllabic words in the Nagamese language could also be broadly sub-grouped into two sub-types. These are:
\begin{enumerate}
\item $V(C)(C)CV(C)(C)CV(C)$
\item $(C)CV(C)(C)V(C)(C)(C)V(C)$
\end{enumerate}

There are few words in the Nagamese language that have tetra syllables. The syllabic structure of the tetrasyllabic words in this language could be schematically presented as follows: 
\begin{center}$(C)V(C)CVCV(C)CV(C)$ \end{center}

There are no pentasyllabic words in the Nagamese language unless one takes clear compound words.

\subsection{Grammar} 
The detailed grammar of Nagamese can be found in the works of Sreedhar \cite{sreedhar1985standardized}, Baishya \cite{baishya2004structure} and Bhattacharjya \cite{bhattacharjya2001genesis}. Some example sentences in Nagamese are given below:-

\begin{enumerate}
\item Sualitu gor bitorte ase (the girl is inside the house)
\item Moy dos baje pora yeti ase (I am waiting here from 10 o'clock)
\item Syama joldi kitab porise (Shyama read the book quickly)
\end{enumerate}

\section{Related Works}
\label{section:related_works}
Since Nagamese is an Assamese-lexified creole language, some of the works done in the Assamese Language are presented here.

Saharia et al.\cite{saharia2009part} worked on developing a part of speech (POS) tagger for Assamese using the Hidden Markov Model (HMM), in which a tagset of 172 tags was developed and performed morphological analysis to determine the probable tags for the unknown words. On a manually tagged corpus of 10k words for training, obtaining an accuracy of 87\% on the test data.

Pathak et al.\cite{pathak2022aspos} worked on a Deep Learning (DL)-based POS tagger for Assamese, using several pre-trained word embeddings, and was able to attain an F1 score of 86.52\%.

Phukan et al.\cite{phukan2024exploring} applied character-level Long Short-Term Memory (LSTM) and Bidirectional Long Short-Term Memory (Bi-LSTM) to part-of-speech (POS) tagging for the Assamese language. The annotated dataset uses the LDCIL Assamese tagset and contains 60,000 words. The LSTM model achieved an accuracy of 92.80\%, whereas the BLSTM model achieved an accuracy of 93.36\%.

Pathak et al.\cite{pathak2023part} proposed a BiLSTM-CRF architecture for Assamese POS tagging using a corpus of 404k tokens. They used word embeddings for its implementation, the two top POS tagging models achieving F1 scores of 0.746 and 0.745. Also, a rule-based approach was developed achieving an F1 score of 0.85. Subsequently, the DL-based taggers were integrated with rule-based to achieve an F1 score of 0.925.

Deka et al.\cite{deka2020study} compared tagging performances of Conditional Random Field and Trigrams'nTag for POS tagging for Assamese Language.

Phukan et al.\cite{phukan2023parts} explored long short-term memory (LSTM) and bidirectional long short-term memory (Bi-LSTM) for POS tagging for Assamese, on a corpus comprising 50k words, achieving an accuracy of 91.20\% for LSTM and 91.72\% for BiLSTM.

Talukdar et al.\cite{talukdar2024deep} used Recurrent Neural Network (RNN) and Gated Recurrent Unit (GRU) to develop a POS tagger for Assamese. The tagset used the Universal Parts of Speech (UPoS) tagset on a dataset of 30k words, achieving F1 scores of 94.01 and 94.56 for RNN and GRU.

Phukan et al.\cite{phukan2023parts} explored the Viterbi algorithm for POS tagging for Assamese, focusing on tagging out-of-vocabulary words. The system used a corpus of 50k words to train the algorithm, achieving an accuracy of 86.34\%.

Talukdar et al.\cite{talukdar2024deep} applied DL models such as GRU, RNN, and Bidirectional LSTM (BiLSTM) architectures for POS tagging for Assamese religious texts, on a dataset of approximately 11,000 sentences.

\section{POS Tagset}
\label{section:pos_tagset}
From the original 36 tags employed in the Penn Treebank Tagset, the tags for the Nagamese POS tagger were reduced to 14 Tags. 
To differentiate Foreign words from Nagamese words a 'FW' tag was introduced into the tag set to be used for the Nagamese tagger. The tagset is shown in Table~\ref{table:pos_tagset}.

\begin{table}[!htb]
  \begin{center}
    \caption{POS Tagset}
    \label{table:pos_tagset}
    \vspace{0.2cm}
    \begin{tabular}{|c|c|c|}
      \hline
      Sl. no. & category  &  tag \\
      \hline
1&	Adjective&	ADJ\\
2&	Adverb&	ADV\\
3&	Conjunction&	CONJ\\
4&	Complementizers&	CMP\\
5&	Determinant&	DET\\
6&	Postposition/Preposition&	PP\\
7&	Interjection&	INTJ\\
8&	Noun&	N\\
9&	Pronoun&	PN\\
10&	Quantifier&	QN\\
11&	Verb&	V\\
12&	Foreign Word&	FW\\
13&	Symbol&	SYM\\
14&	Unknown&	UNK\\
15&	Numeral&	NUM\\
      \hline
    \end{tabular}
  \end{center}
\end{table}

\section{Methodology}
\label{section:methodology}
\subsection{Dataset Creation}
\begin{figure*}
	\centering
	\[\includegraphics[scale=0.15]{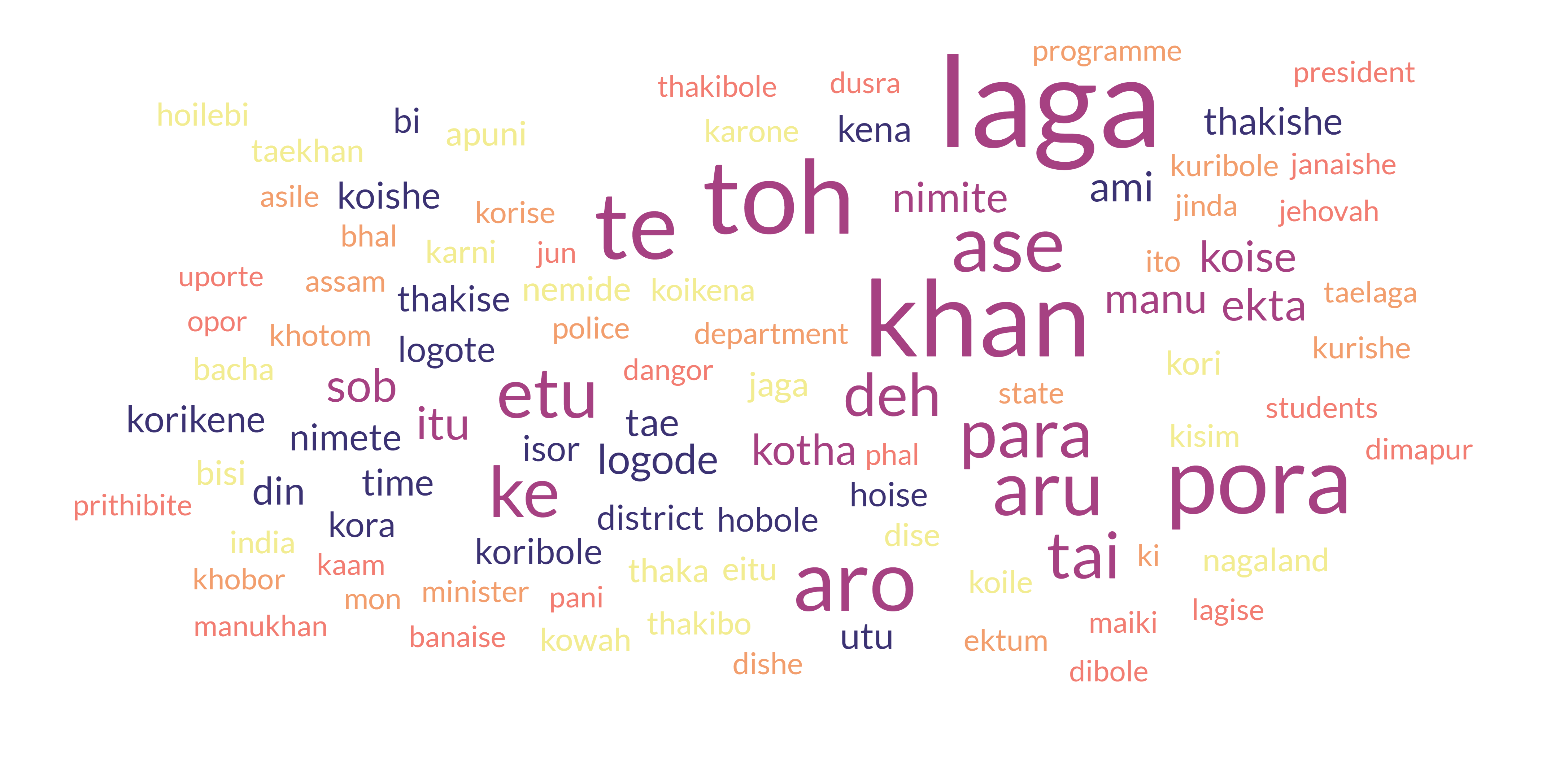} \]
	\caption{Nagamese wordcloud based on word frequency}
	\label{fig:nagamese_wordcloud}
\end{figure*}

The Nagamese  Corpus, which contains approximately 26,000 words, was created by collecting articles from a local newspaper, 'Nagamese Khobor'. The 'Nagamese khobor' newspaper contains a variety of content related to current state affairs, sports, etc. Based on the word frequency, a word cloud is shown in Fig.~\ref{fig:nagamese_wordcloud}.

Random publications of the Nagamese newspaper and bible phrases were collected, from which various articles were extracted to obtain a mixed corpus. The corpus was manually annotated by one annotator who is a native speaker of Nagamese. We manually annotated a corpus of 16,115 tokens, the tag frequencies shown in Fig.~\ref{table:tag_frequencies}. To validate the dataset, another annotator was employed to manually tag 1,864 tokens. Out of these, 125 tokens were disagreed and 102 were Foreign words. Hence, including foreign words the disagreement is 6.7\%, and excluding these foreign words, the disagreement is 1.23\%.

Sample of the annotated dataset is shown below:-\\
Titia/ADV Isor/N koise/V ,/SYM “/SYM Ujala/N hobole/V dibi/V ./SYM ”/SYM Aru/CONJ Ujala/N hoise/V ./SYM\\
Itu/ADJ dikhikena/V Isor/N khusi/ADJ lagise/V ./SYM

\begin{table}[!htb]
  \begin{center}
    \caption{Tag frequencies}
    \label{table:tag_frequencies}
    \vspace{0.2cm}
    \begin{tabular}{|c|c|c|}
      \hline
      Sl. no. &  tag & frequency\\
      \hline
1&	ADJ & 1507\\
2&	ADV & 709\\
3&	CONJ & 591\\
4&	CMP & 35\\
5&	DET & 132\\
6&	PP & 2418\\
7&	INTJ & 65\\
8&	N & 1804\\
9&	PN & 1141\\
10&	QN & 84\\
11&	V & 1678\\
12&	FW & 3744\\
13&	SYM & 1830\\
14&	UNK & 143\\
15&	NUM & 234\\
      \hline
    \end{tabular}
  \end{center}
\end{table}

\subsection{Conditional Random fields Model}

\begin{figure}[!htbp] 
	\centering
	\[\includegraphics[scale=0.3]{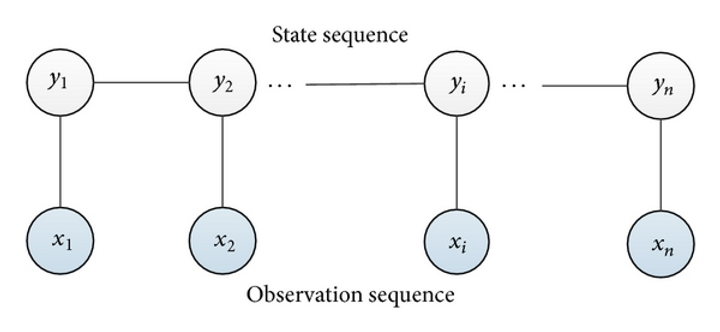} \]
	\caption{Linear-chain CRF model}
	\label{fig:crf_model}
\end{figure}

Conditional random fields (CRFs) fall into the sequence modeling family and are a class of statistical modeling methods often applied in pattern recognition and machine learning and used for structured prediction. Whereas a discrete classifier predicts a label for a single sample without considering "neighboring" samples, a CRF can take context into account; e.g., the neighboring tokens information in a sequence.
CRFs are a type of discriminative undirected probabilistic graphical model. It overcomes the problem in Maximum Entropy Markov Models (MEMM) known as the "Label bias" problem. Fig.~\ref{fig:crf_model} shows the linear chain CRF model.

We have implemented CRF using the sklearn-crfsuite~\footnote{https://sklearn-crfsuite.readthedocs.io/en/latest/} library.

The features used are:-\\
-the current word\\
-whether it is the first word in the sentence\\
-whether it is the last word in the sentence\\
-whether the word is capitalized\\
-whether the word is in lowercase\\
-prefix details up to length 3\\
-suffix details up to length 3\\
-previous word\\
-next word\\
-contains hyphen\\
-is numeric\\
-contains upper case inside word

A sample of the generated features is given below:-\\
'has\_hyphen': False, 'is\_first': True, 'suffix-4': 'itia', 'is\_numeric': False, 'prefix-3': 'Tit', 'word': 'Titia', 'suffix-1': 'a', 'is\_capitalized': True, 'next\_word': 'Isor', 'prefix-1': 'T', 'is\_all\_caps': False, 'prev\_word': '', 'is\_all\_lower': False, 'is\_last': False, 'suffix-2': 'ia', 'suffix-3': 'tia', 'capitals\_inside': False, 'prefix-2': 'Ti',....

For Gradient descent, the L-BFGS method has been used, 100 iterations have been used for the optimization algorithm, and to avoid the overfitting problem, L1 and L2 regularizers have been employed.

\section{Results and Discussions}
\label{section: experimental_results}
\begin{figure*}[!ht] 
	\centering
	\[\includegraphics[scale=0.3]{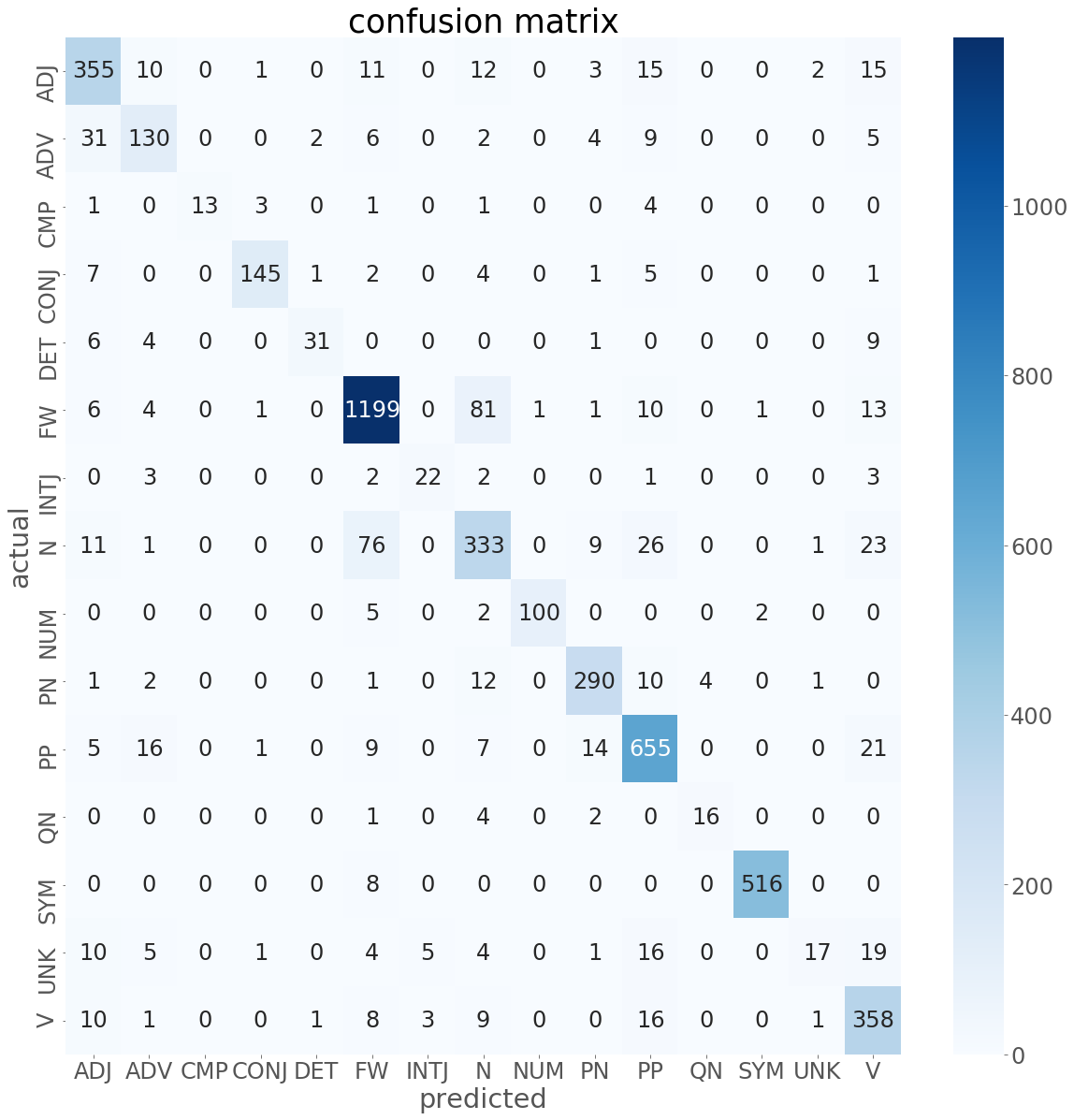} \]
	\caption{Confusion matrix for POS tagging}
	\label{fig:confusion_matrix}
\end{figure*}

This section presents the results and discusses the results of the tagging. The annotated dataset comprises a total of 16,115 tokens (749 sentences). The training: test split used is 70:30\%. Table~\ref{tagging_results} reports the results of the POS tagging. We obtain an overall tagging accuracy of 85.70\%; precision, and recall of 86\%, and f1-score of 85\%. An error analysis for each tagset in the POS tagset is reported in the form of a confusion matrix as shown in Fig.~\ref{fig:confusion_matrix}.

The accuracy measures used for the model's performance are precision, recall, and f1-score.\\

\underline{Precision}\\
Precision measures how many of the positive predictions made by the model are actually correct, expressed mathematically as:

Precision = \[\frac{True\_Positive}{(True\_Positive + False\_Positive)}\]

\underline{Recall}\\
Recall measures the completeness of positive predictions, expressed mathematically as:

Recall = \[\frac{True\_Positive}{(True\_Positive + False\_Negative)}\]

\underline{F1-Score}\\
F1-Score is the harmonic mean between recall and precision and tells us how precise and robust our classifier is, expressed mathematically as: 

F1-Score =  \[\frac{2 * (Precision* Recall)}{(Precision + Recall)}\]

\begin{table}[!htb]
  \begin{center}
    \caption{POS tagging results}
    \label{tagging_results}
    \vspace{0.2cm}
    \footnotesize
    \begin{tabular}{|c|c|c|c|c|}
      \hline
      Tag & precision  & recall & f1-score  & support \\
      \hline
      \hline

        ADJ   &    0.80  &    0.84  &    0.82   &    424\\
        ADV    &   0.74   &   0.69   &   0.71    &   189\\
        CMP    &   1.00   &   0.57   &   0.72    &    23\\
       CONJ    &   0.95   &   0.87   &   0.91    &   166\\
        DET    &   0.89   &   0.61   &   0.72    &    51\\
         FW    &   0.90   &   0.91   &   0.90    &  1317\\
       INTJ    &   0.73   &   0.67   &   0.70    &    33\\
          N    &   0.70   &   0.69   &   0.70    &   480\\
        NUM    &   0.99   &   0.92   &   0.95    &   109\\
         PN    &   0.89   &   0.90   &   0.90    &   321\\
         PP    &   0.85   &   0.90   &   0.88    &   728\\
         QN    &   0.80   &   0.70   &   0.74    &    23\\
        SYM    &   0.99   &   0.98   &   0.99    &   524\\
        UNK    &   0.77   &   0.21   &   0.33    &    82\\
          V    &   0.77   &   0.88   &   0.82    &   407\\

avg / total   &    0.86   &   0.86   &   0.85  &    4877\\

      \hline
    \end{tabular}
  \end{center}
\end{table}

From the experiment conducted as shown in Table~\ref{tagging_results}, we obtained a precision of 1.0 for CMP. The lowest precision for N with 0.70. The highest recall for SYM with 0.98 and the lowest recall for UNK with 0.21. The highest f1-score for 0.99 for SYM, and the lowest f1-score for UNK with 0.33.

The misclassified cases are listed below:-\\
i. ADJ tagged as ADV (10), CONJ (1), FW (11), N (12), PN (3), PP (15), UNK (2), and V (15).\\
ii. ADV tagged as ADJ (31), DET (2), FW (6), N (2),
PN (4), PP (9), and V (5).\\
iii. CMP tagged as ADJ (1), FW (1), N(1), and CONJ (3).\\
iv. CONJ tagged as ADJ (7), DET (1), FW (2), N (4), PN (1), PP (5), and V (1).\\
v. DET tagged as ADJ (6), ADV (4), PN (1), and V (9).\\
vi. FW tagged as ADJ (6), ADV (4), CONJ (1), N (81), NUM (1), PN (1), PP (10), SYM (1), and V (13).\\
vii. INTJ tagged as ADV (3), FW (2), N (2), PP(1), and V (3).\\
viii. N tagged as ADJ (11), ADV (1), FW (76), PN (9), PP(26), UNK (1), and V (23).\\
ix. NUM tagged as FW (5), N (2), and SYM (2).\\
x. PN tagged as ADJ (1), ADV (2), FW (1), N (12), PP(10), QN (4), and UNK (1).\\
xi. PP tagged as ADJ (5), ADV (16), CONJ (1), FW (9), N (7), PN(14), and V (21).\\
xii. QN tagged as FW (1), N (4), PN (2), QN (16), and UNK (17).\\
xiii. SYM tagged as FW (8).\\
xiv. UNK tagged as ADJ (10), ADV (5), CONJ (1), FW (4), INTJ (5), N (4), PN (1), PP (16), and V (19).\\
xv. V tagged as ADJ (10), ADV (1), DET (1), FW (8), INTJ (3), N (9), PP (16), and UNK (1).\\
Note: The figure in bracket indicates the number of tokens.

\begin{figure}[!htbp] 
	\centering
	\[\includegraphics[scale=0.8]{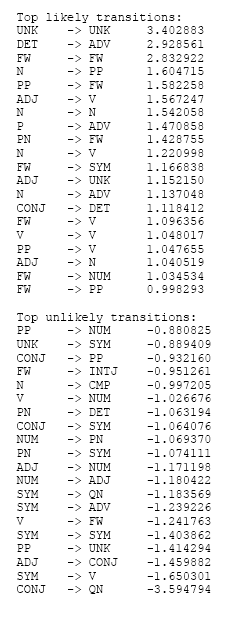} \]
	\caption{Top likely and unlikely transitions}
	\label{fig:transitions}
\end{figure}

Fig.~\ref{fig:transitions} shows the top likely and unlikely transitions from one tag to the other tag. The top likely transition is from UNK to UNK and the top unlikely transition is from PP to NUM.

\section{Conclusion}
\label{section:conclusion}
In this paper, a part-of-speech tagger has been built for Nagamese Language using machine learning techniques by creating a POS-tagged Nagamese corpora of size 16,115 tokens. Using Conditional Random Fields (CRF), an overall tagging accuracy of 85.70\%; precision, and recall of 86\%, and an f1-score of 85\% is obtained. The main contribution of this work is the creation of the POS annotated corpus and the evaluation of the annotated corpus. Some of the limitations of this work are the number of tags used and the size of the dataset.

Future works include:- i) Increasing the number of tags in the tagset, ii) Increasing the size of the tagged corpus, iii) Using the developed POS tagger to build other applications such as sentiment analysis, machine translation, etc for the Nagamese Language, and iv) exploring transfer learning from the Assamese Language, etc.

\section*{Acknowledgements} 
We would like to thank the Department of Information Technology, Nagaland University for providing research facility to conduct our experiments.

\small{
\bibliographystyle{cys}
\bibliography{biblio}
}
\normalsize

\begin{biography}[]{} 
\end{biography}

\end{document}